\documentclass[conference]{IEEEtran}
\IEEEoverridecommandlockouts
\usepackage{amsmath,amssymb,amsfonts}
\usepackage{algorithmic}
\usepackage{graphicx}
\usepackage{textcomp}
\usepackage{xcolor}
\usepackage{multirow}
\usepackage{csquotes}
\usepackage{xurl}

\def\BibTeX{{\rm B\kern-.05em{\sc i\kern-.025em b}\kern-.08em
    T\kern-.1667em\lower.7ex\hbox{E}\kern-.125emX}}
\begin{document}

\title{Anomaly detection in surveillance videos using transformer based attention model}

\author{\IEEEauthorblockN{Kapil Deshpande}
\IEEEauthorblockA{\textit{Dept. of Information Technology} \\
\textit{Indian Institute of Information Technology Allahabad}\\
Prayagraj, Uttar Pradesh, India \\
mit2020040@iiita.ac.in}
\\
\IEEEauthorblockN{Sanjay Kumar Sonbhadra}
\IEEEauthorblockA{\textit{Dept. of Computer Science and Engineering} \\
\textit{ITER, Shiksha ‘O’ Anusandhan}\\
Bhubaneswar, Odisha, India \\
sanjaykumarsonbhadra@soa.ac.in}
\and
\IEEEauthorblockN{Narinder Singh Punn}
\IEEEauthorblockA{\textit{Dept. of Information Technology} \\
\textit{Indian Institute of Information Technology Allahabad}\\
Prayagraj, Uttar Pradesh, India \\
pse2017002@iiita.ac.in}
\\
\IEEEauthorblockN{Sonali Agarwal}
\IEEEauthorblockA{\textit{Dept. of Information Technology} \\
\textit{Indian Institute of Information Technology Allahabad}\\
Prayagraj, Uttar Pradesh, India \\
sonali@iiita.ac.in}
}

\maketitle

\begin{abstract}

Surveillance footage can catch a wide range of realistic anomalies. This research suggests using a weakly supervised strategy to avoid annotating anomalous segments in training videos, which is time consuming. In this approach only video level labels are used to obtain frame level anomaly scores. Weakly supervised video anomaly detection (WSVAD) suffers from the wrong identification of abnormal and normal instances during the training process. Therefore it is important to extract better quality features from the available videos. With this motivation, the present paper uses better quality transformer-based features named Videoswin Features followed by the attention layer based on dilated convolution and self attention to capture  long and short range dependencies in temporal domain. This gives us a better understanding of available videos. The proposed framework is validated on real-world dataset i.e. ShanghaiTech Campus dataset which results in competitive performance than current state-of-the-art methods. The model and the code are available at \url{https://github.com/kapildeshpande/Anomaly-Detection-in-Surveillance-Videos}.

\end{abstract}


\section{Introduction}
Video anomaly detection has gained a lot of attention due to its applications in surveillance systems. The cost of deploying surveillance systems has reduced significantly in recent years but it still requires human intervention in detecting anomalous events like fighting, abusing, stealing, etc. Considering the additional cost of human labor and the loss of productive time, the development of intelligent algorithms for video anomaly detection is required. The vague nature of anomaly and the unavailability of annotated data makes anomaly detection difficult. There are various unsupervised \cite{b1,b2} and weakly supervised \cite{b11,b12,b13} solutions present. Generally, unsupervised anomaly detection tries to learn the distribution of normal events and mark outliers as anomalies. Since it is impossible to learn all possible normal events distribution therefore this model is highly biased and fails in the case of real-world events. It tries to combat the above problem by using both normal and anomalous events while training. 

Weakly supervised approaches require significantly less effort as compared to supervised learning because it requires only video-level labels instead of frame-level. However the major challenge of weakly supervised approaches is in identifying the abnormal snippets from the anomalous videos, this is because: the abnormal videos contain a large number of normal snippets and the abnormal events can have only slight differences from normal events. All these above issues can be resolved by using multiple instance learning (MIL) \cite{b5} where the training set is divided into the same numbers of abnormal and normal snippets. Two bags are created i.e.  a normal bag which contains snippets from a normal video and an abnormal bag that contains snippets from an abnormal video. The snippet with the maximum anomaly score is selected from each bag and the loss is back propagated. Although this method partially addresses the previous issues but it also introduces the following problems: The highest anomaly score can be from normal bag instead of abnormal bag and when anomaly videos have multiple anomaly in the single video it fails to leverage additional anomalies because it only take the snippet with highest anomaly score from the abnormal bag.

To address all these issues the proposed solution used the Robust Temporal Feature Magnitude (RTFM) learning model inspired by Tian et al. \cite{b7}. This model relies on the temporal feature magnitude i.e. l2 norm of features for anomaly detection, where normal snippets are represented by low magnitude features, while abnormal snippets are represented by high magnitude features. This model try to maximize $\delta_{\mathrm{score}}$ which denotes the difference between the mean of l2 norm of top K features from the abnormal and normal bag where K is the number of abnormal snippets in an abnormal video. This solves the previously discussed problem because: It's more likely to choose anomalous snippets from abnormal videos instead of normal videos. It can utilize multiple anomalies in the anomaly videos which will result in better utilization of training data.

It is important to extract better features from the available videos to avoid the wrong identification of abnormal and normal instances during the training process. Researchers have been inspired to employ video transformers as feature extractors to handle anomaly detection tasks as a result of their recent success with video classification tasks. Therefore the proposed model uses a transformer based features named Videoswin Features \cite{b40} which have consistently outperformed the CNN based models like I3D \cite{b39}, C3D \cite{b38}, etc. The feature extraction is followed by an attention layer based on dilated convolution to capture most relevant long and short range dependencies \cite{b44,b45,b46,b47}. The proposed solution is validated on a real-world dataset i.e. ShanghaiTech Campus dataset which results in competitive performance than current state-of-the-art methods. The major contribution of the present research work are described below:
\begin{itemize}
    \item To improve the understanding of given videos, a newer transformer based feature extraction model is used named videoswin transformer.
    \item To highlight relevant features an attention layer based on dilated convolution and self attention that captures long and short range temporal dependencies.
    \item A comparative study with current state-of-the-art approaches is conducted to examine the effects of the proposed model on the open source ShanghaiTech dataset. The proposed model achieved competitive performance (AUC score) than current state-of-the-art methods.
\end{itemize}

The rest of the paper is divided into various sections, where Section 2 covers the prevailing work in anomaly detection, followed by proposed methodology in Section 3. Section 4 presents the experimental analysis and finally concluding remarks are presented in Section 5.

\section{Related Work}
Traditional video anomaly detection uses unsupervised learning \cite{b3,b4} algorithms where it tries to learn the distribution of normal events and mark outliers as anomalies. Since it is impossible to learn all possible normal events distribution therefore this model is highly biased and fails in the case of real-world events. Other methods use one-class classification \cite{b1,b2} assuming only normal labeled data is available. Some approaches rely on tracking \cite{b25,b26} to model people's regular movement and identify deviations as anomalies. Since it is tough to acquire accurate information of tracks, numerous strategies for avoiding tracking and learning global motion patterns, such as topic modeling \cite{b28}, have been used, context-driven method \cite{b32} social force models \cite{b30}, histogram-based methods \cite{b27}, motion patterns \cite{b29}, Hidden Markov Model (HMM) on local spatiotemporal volumes\cite{b31}, and mixtures of dynamic textures model \cite{b32}. These techniques learn distributions of normal motion patterns from training videos of normal behaviors and discover low likely patterns as anomalies. After the initial success of the sparse representation and dictionary learning methodologies, researchers employed sparse representation \cite{b34,b35} to learn the dictionary of patterns. Where anomalous patterns have high reconstruction errors during testing. After the initial success of deep learning in image classification, many techniques for video action classification \cite{b36,b37} have been developed.

Alternatively, some approaches rely on data reconstruction utilizing generative models to learn normal sample representations by (adversarial) reducing the reconstruction error \cite{b19,b20,b21}. These methods presume that undetected abnormal videos/images can often be poorly reconstructed, and samples with large reconstruction errors are considered anomalies. These techniques may overfit the training data and fail to distinguish abnormal from normal events due to a lack of prior knowledge about anomaly.

To solve the above problems Sultani et al. \cite{b5} introduced a weakly supervised solution that can learn anomaly patterns by using both normal and anomaly videos using MIL-based models with CNN as the backbone for feature extraction. However, it fails to separate noise present in the positive bag and this can lead to normal snippets being mistaken as abnormal. In this context, to clear the noise present in the positive bag Zhong et al. \cite{b6} proposed to use a graph convolution neural network. Although it partially solved the problem, it was computationally heavy. The RTFM model \cite{b7} which solves the above problem by using the l2 norm-based ranking loss function. Although it still uses CNN as the backbone for feature extraction.

To capture consistency between successive frames, traditional attention approaches employ consecutive frames and transform them into handcrafted motion trajectories. Other methods such as stacked RNN \cite{b16}, LSTM \cite{b11}, convolutional LSTM \cite{b11}, and GCN-based \cite{b6} can capture short-range fixed-order temporal dependencies but they either fail to capture long-range dependencies or they are computationally expensive. Following this context, the proposed attention layer uses dilated convolution based attention mechanism which captures short and long-range temporal dependencies and is computationally inexpensive as compared to other methods. The proposed method uses a more effective transformer-based model for feature extraction and a temporal attention layer for necessary feature enhancement, thus improving the model's overall performance.

\section{Proposed Method}
\begin{figure}[]
\centerline{\includegraphics[width=\columnwidth]{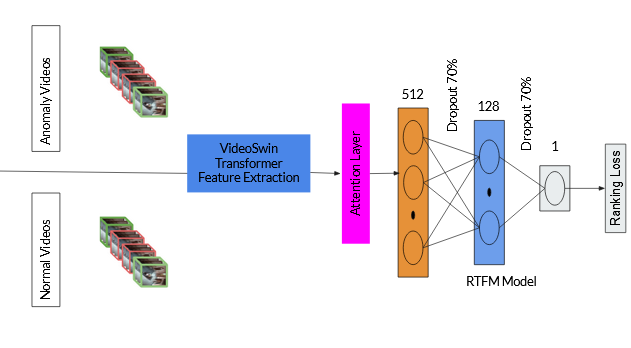}}
\caption{The proposed model architecture.}
\label{fig1}
\end{figure}

The proposed framework is divided into 3 stages as given in Fig.~\ref{fig1}. As the exact location or frame-level labels are not provided for learning, the proposed solution follows a weakly supervised learning where the videos of different duration are divided into a fixed number of snippets containing the same number of frames. The proposed solution assumed that snippets obtained from an anomaly video contain at least one anomaly snippet, but the snippets from normal videos contain all normal snippets. In stage 1, A pre-trained videoswin model for feature extraction of the snippets. In stage 2 an attention layer is applied to the feature to capture relevant  long and short range dependencies in the temporal domain. At last in stage 3 RTFM model is used for anomaly detection on the features obtained from stage 3.

\subsection{Stage 1 (Feature Extraction)}
To extract features this paper uses videoswin transformer model which is trained on large-scale datasets like Kinetics \cite{b41} and ImageNet \cite{b42}. The use of a pretrained model allows us to extract better quality features. Traditional transformer models calculate self-attention with respect to all the elements present but in the case of images, it is computationally expensive to perform. To solve this issue, the swin transformer divides images into windows and calculates self-attention inside this window only. Now it slides the window on the images to get the self-attention value of the whole set of images more efficiently.

To begin the feature extraction process, the first step is to divide the videos into frames (of size let's say H $\times$ W). Now the set of these frames (T) makes the video for feature extraction where each video has RGB channels. This gives us the input dimension as N $\times$ C $\times$ T $\times$ H $\times$ W, where N is the batch size and C is the number of channels.

\subsection{Stage 2 (Attention Layer)} 
The main objective of this stage is to learn the discriminative representation of normal and abnormal snippets by improving the quality of the feature map obtained from stage 1. This objective is achieved using an attention layer that can encode the long and short range dependencies in temporal domain on the feature map while drawing focus of the model towards most relevant features.

\begin{figure}[]
\centerline{\includegraphics[width=\columnwidth]{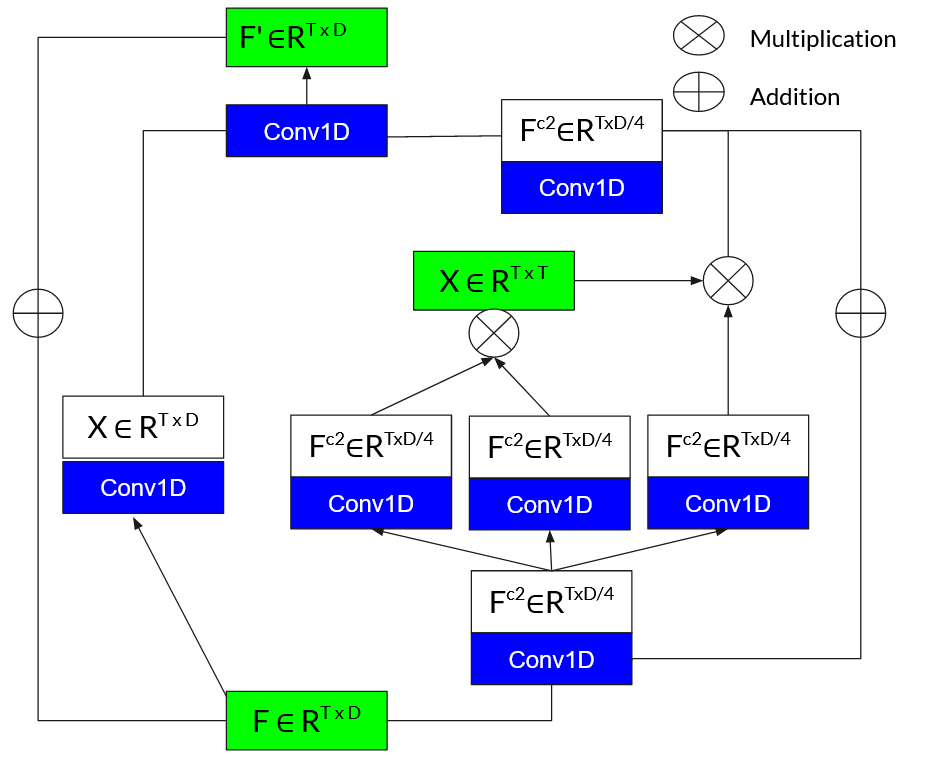}}
\caption{The proposed attention layer architecture}
\label{fig2}
\end{figure}

The proposed attention layer is shown in Fig.~\ref{fig2}. Given an input feature map $F\in\mathbb{R}^{T\times D}$, it produces the output attention feature maps $F'\in\mathbb{R}^{T\times D}$. It consists of two modules, the one on the left is a short range module, it is used to capture short-term temporal dependencies and the one on the right is a long range module it is used to compute global temporal context.

To calculate the global temporal context, the pairwise temporal self attention is calculated which produces the feature map  $M\in\mathbb{R}^{T\times T}$. It first applies the conv1D layer to reduce information to $F^{c}\in\mathbb{R}^{T\times D/4}$ where $F^{c} = \mathrm{conv1D}(F)$, then it applies 3 conv1d layers separately. $F^{c1} = \mathrm{conv1D}(F^{c})$, $F^{c2} = \mathrm{conv1D}(F^{c})$, $F^{c3} = \mathrm{conv1D}(F^{c})$. It will combine these 3 conv1D layers with $F^{c4} = \mathrm{conv1D}( (F^{c1} * (F^{c2})^{T} ) * F^{c3})$. A residual is added, which gives the final output, $M = F^{c4} + F^{c}$ , where  $M\in\mathbb{R}^{T\times T}$.

To calculate the short term temporal dependencies it applies the conv1D layer which gives it the output, $K = \mathrm{conv1D}(F)$, where $F\in\mathbb{R}^{T\times D}$. The output $M$ from the long range module is concatenated with the output $K$ from the short range module and a residual connection is added to give us the final output, $F'=\mathrm{concat}(M, K) + F^{c}$, where  $F\in\mathbb{R}^{T\times D}$.

\subsection{Stage 3 (Anomaly Detection)}
The proposed anomaly detection model uses Robust Temporal Feature Magnitude Learning (RTFM) model, in which temporal feature magnitude i.e. l2 norm of video snippets are used for anomaly detection where normal snippets are represented by low magnitude features, while abnormal snippets are represented by high magnitude features. The proposed model assumes that anomalous snippets have a larger mean feature magnitude than normal snippets.

Let $\|x\|$ be the feature magnitude of snippets where $x^+$ means abnormal snippet and $x^-$ means normal snippet, which are obtained by normal ($X^+$) and abnormal ($X^-$) videos. Model learns by trying to maximize the $\delta_{\mathrm{score}}(X^+, X^-)$ which denotes the difference between the mean of l2 norm of topK features from the abnormal and normal bag where k is the number of abnormal snippets in abnormal video. To maximize the $\delta_{\mathrm{score}}(X^+, X^-)$, the loss function (shown in \ref{eqn:eq1}) is optimized during backpropagation.
\begin{equation}
\label{eqn:eq1}
\begin{aligned}
L(X^+, X^-) = max(0, m - mean(topK(\|X^+\|)) \\+ mean(topK(\|X^-\|))
\end{aligned}
\end{equation}
where $m$ is a constant predefined margin.

A binary cross-entropy based loss function is applied to learn the snippet classifier as shown in \ref{eqn:eq2}. It trains a snippet classifier with 0 and 1 class labels indicating normal and abnormal snippets respectively.
\begin{equation}
\label{eqn:eq2}
loss = - ylog(x) + (1 - y) log(1-x)
\end{equation}
where $x$ is the mean of l2 norm of topK features, $x = mean(topK(\|X^+\|))$, and $y$ is the binary value indicating actual class labels as normal or abnormal.
\begin{equation}
\label{eqn:eq3}
Smoothness = \Sigma  f(v^{i}) - f(v^{i+1})^{2}
\end{equation}

Temporal smoothness is used between consecutive video snippets to vary anomaly score smoothly between video snippets.
\begin{equation}
\label{eqn:eq4}
Sparsity =  \Sigma f(v\textsubscript{i})
\end{equation}

Anomaly frequently happens over a brief period of time in real-world circumstances which leads to sparse anomaly scores of segments in the anomalous bag. To avoid this issue, a sparsity term is used.
\begin{equation}
\label{eqn:eq5}
\begin{aligned}
Final loss = \lambda1 * \mathrm{Eq.} \ref{eqn:eq1} + \lambda2 * \mathrm{Eq.} \ref{eqn:eq2} + \\ \lambda3 * \mathrm{Eq.} \ref{eqn:eq3} + \lambda4 * \mathrm{Eq.} \ref{eqn:eq4} 
\end{aligned}
\end{equation}
where $\lambda$'s are the respective learning rates for the Eq.s.

\section{Experiments}
\subsection{Dataset Description} 
\begin{figure}[h]
\centerline{\includegraphics[width=\columnwidth]{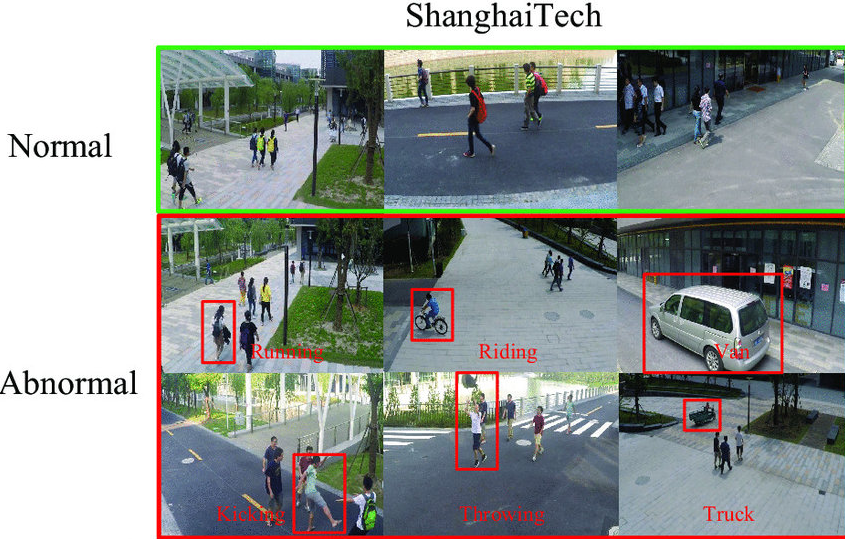}}
\caption{ShanghaiTech dataset normal and abnormal clips}
\label{fig3}
\end{figure}

This paper uses a large-scale video anomaly detection dataset called the ShanghaiTech Campus dataset \cite{b17}. It includes video from fixed angle street surveillance cameras. It has 437 videos from 12 different backgrounds, with 130 anomalous and 307 normal videos. This is a popular benchmark dataset for anomaly detection tasks that uses both anomalous and normal data. To restructure the dataset into a weakly supervised training set, Zhong et al. \cite{b6} picked a sample of anomalous testing videos and turned them into training videos so that all 13 background scenes are covered by the training and testing set. To convert the dataset into weekly supervised, this paper used the same approach as used by Zhong et al. \cite{b6} and Tian et al. \cite{b7}. Fig. \ref{fig3} shows the sample normal and abnormal clips from the dataset.

\subsection{Evaluation Metric}
To measure the model's performance, this paper used frame-level receiver operating characteristics (ROC) as well as its area under the curve (AUC) score, following the previous methods \cite{b5,b6,b7}. AUC score is a measure of separability. It represents the model's ability to discriminate between classes. An AUC score of 1 means the model can separate both classes perfectly. AUC score of 0 means the model is reciprocating the results means it predicts positive class as negative and vice-versa. AUC score of 0.5 means the model has no class separation capacity. The ROC curve is plotted at all possible classification thresholds by calculating the values of TPR and FPR at every threshold from 0 to 1.

\subsection{Implementation Details}
For feature extraction from pre-trained videoswin transformer model on Kinetics dataset, each video is divided into frames of size 224 x 224, for video is divided into T = 32 temporal segments where each segment is 16 frames long, this gives us ten crop features of dimensions 32 x 1024. Cropping snippets into the four corners, center and their flipped form is referred as ten cropping.

In Eq. \ref{eqn:eq1} the margin, $m = 100$ and the value of $k = 3$. The 3 fully connected (FC) layers in the RTFM model have 512, 128 and 1 nodes respectively where $1^{st}$ and $2^{nd}$ FC layer is followed by a ReLU activation function and the last layer is followed by sigmoid function. A dropout function is added after every layer with rate = 0.7. The model is trained using adam optimizer with weight decay of 0.005 and learning rate of 0.001 with batch size = 32 for 500 epochs. Each mini batch has 32 samples chosen at random from normal and abnormal videos. For fare comparison, this paper used the same benchmark setup used by Sultani et al.\cite{b5}, Zhong et al. \cite{b6} and Tian et al. \cite{b7}.

\subsection{Result Analysis}
\begin{figure}[]
\centerline{\includegraphics[width=0.6\columnwidth]{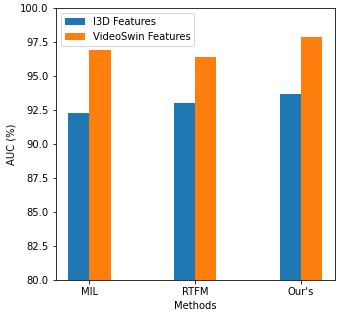}}
\caption{Comparison I3D vs Videoswin Features}
\label{fig10}
\end{figure}

\begin{figure}[]
\centerline{\includegraphics[width=\columnwidth]{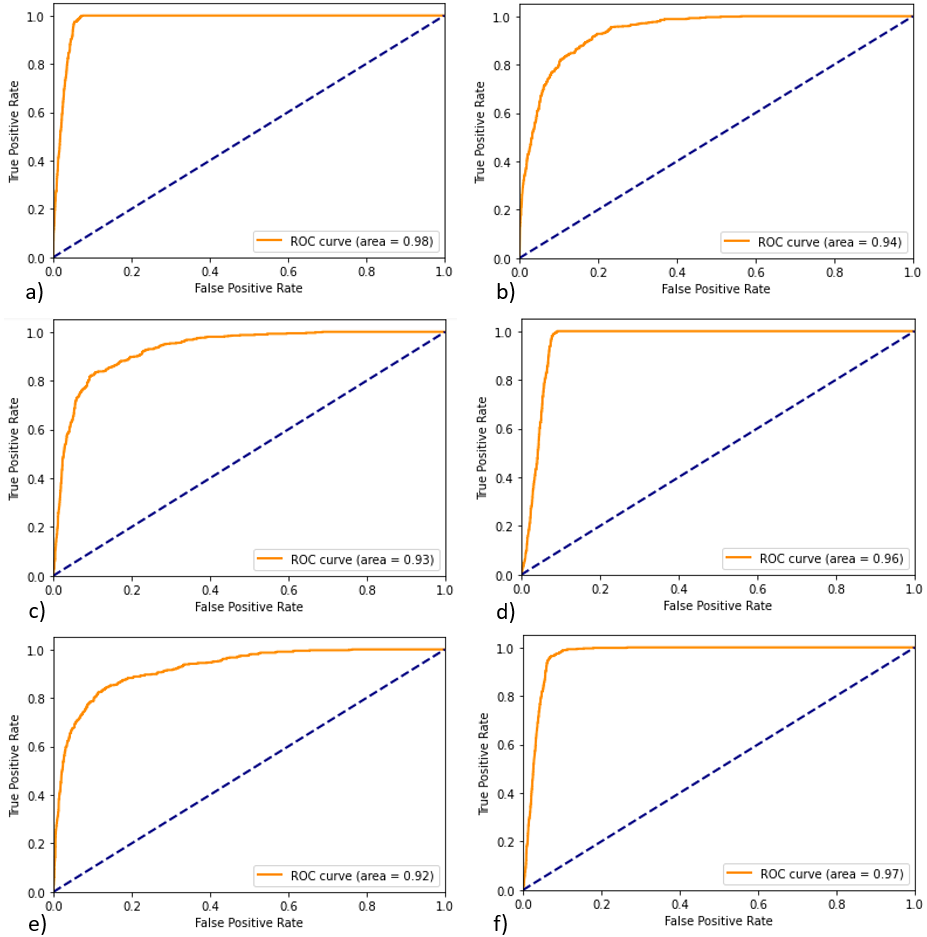}}
\caption{ROC curves of the a) proposed model using videoswin backbone, b) proposed model using I3D backbone, c) RTFM model using I3D backbone, d) RTFM model using videoswin backbone, e) MIL model using I3D backbone, and f) MIL model using videoswin backbone}
\label{fig4}
\end{figure}

The results are reported on the ShanghaiTech Campus dataset \cite{b17}. Where two backbone models for feature extraction are used namely I3D \cite{b39} and videoswin \cite{b40}. Comparisons with the previous weakly supervised solutions are given in Table \ref{tab1} and visually presented in Fig. \ref{fig10}. Furthermore, the inference drawn is analysed with the help of ROC curves as given in Fig. \ref{fig4}. 

\begin{table}[!b]
\caption{The comparative analysis of video anomaly detection models. The best outcomes are shown in bold font.}
\label{tab1}
\begin{center}
\begin{tabular}{|l|l|l|}
\hline
\textbf{Method} &  \textbf{Feature} & \textbf{AUC}\\
\hline
MIL &  {I3D} & 92.3\\
\hline
MIL &  {Videoswin} & 96.9\\
\hline
RTFM & {I3D} & 93.0\\
\hline
RTFM & {Videoswin} & 96.4\\
\hline
Proposed Model & {I3D} & 93.7\\
\hline
\textbf {Proposed Model} & \textbf{{Videoswin}} & \textbf{97.9}\\
\hline
\end{tabular}
\end{center}
\end{table}

The usage of videoswin features leads to better performance than I3D features because of improved video understanding. MIL model specially performed better with videoswin features, it even outperformed the RTFM model which was previously performing better than MIL with I3D features. To compare the attention layer introduced, this paper added a different attention layer to previous methods namely LSTM, CBAM \cite{b8}, RTFM’s Attention Layer \cite{b7}. The results obtained are given in Table \ref{tab2}. With the introduction of the proposed attention layer a better AUC score of around 1\% is acquired. The usage of LSTM and CBAM results in decreased performance because the models fail when high dimensional feature maps are given as input \cite{b43}.

\begin{table}[]
\caption{The comparative analysis of various attention layers on  the video anomaly detection models. The best outcomes are shown in bold font.}
\label{tab2}
\begin{center}
\begin{tabular}{|l|l|l|}
\hline
\textbf{Method} &  \textbf{Feature} & \textbf{AUC}\\
\hline
MIL + LSTM &  {I3D} & 89.0\\
\hline
MIL + LSTM &  {Videoswin} & 96.6\\
\hline
RTFM + LSTM & {I3D} & 89.0\\
\hline
RTFM + LSTM & {Videoswin} & 96.6\\
\hline
MIL + CBAM &  {I3D} & 88.0\\
\hline
MIL + CBAM &  {Videoswin} & 96.9\\
\hline
RTFM + CBAM & {I3D} & 87.5\\
\hline
RTFM + CBAM & {Videoswin} & 96.2\\
\hline
RTFM + No Attention & {I3D} & 91.0\\
\hline
RTFM + No Attention & {Videoswin} & 97.1\\
\hline
\textbf {Proposed Model} & \textbf{{Videoswin}} & \textbf{97.9}\\
\hline
\end{tabular}
\end{center}
\end{table}

\section{Conclusion}
In this research work, a weakly supervised strategy is proposed. It uses better quality features extracted from videoswin transformer model, followed by an attention layer to encode the long and short range dependencies in the temporal domain. The use of the robust temporal feature magnitude (RTFM) model makes this approach better than multiple instance learning (MIL) based techniques because it learns more discriminative features than the MIL model and it exploits abnormal data more easily. It is found from experiments that the use of better quality features and an improved attention layer resulted in improved performance of the model. In future, more experiments can be performed by exploring different strategies to minimize the noise present in the positive bag.

\bibliographystyle{plain}
\bibliography{biblio}

\begin{thebibliography}{10}

\bibitem{b47}
Sonali Agarwal and GN~Pandey.
\newblock Svm based context awareness using body area sensor network for
  pervasive healthcare monitoring.
\newblock In {\em Proceedings of the First International Conference on
  Intelligent Interactive Technologies and Multimedia}, pages 271--278, 2010.

\bibitem{b46}
Prachi Agrawal, Narinder~Singh Punn, Sanjay~Kumar Sonbhadra, and Sonali
  Agarwal.
\newblock Impact of attention on adversarial robustness of image classification
  models.
\newblock In {\em 2021 IEEE International Conference on Big Data (Big Data)},
  pages 3013--3019. IEEE, 2021.

\bibitem{b1}
Arslan Basharat, Alexei Gritai, and Mubarak Shah.
\newblock Learning object motion patterns for anomaly detection and improved
  object detection.
\newblock In {\em 2008 IEEE conference on computer vision and pattern
  recognition}, pages 1--8. IEEE, 2008.

\bibitem{b25}
Arslan Basharat, Alexei Gritai, and Mubarak Shah.
\newblock Learning object motion patterns for anomaly detection and improved
  object detection.
\newblock In {\em 2008 IEEE conference on computer vision and pattern
  recognition}, pages 1--8. IEEE, 2008.

\bibitem{b41}
Joao Carreira, Eric Noland, Chloe Hillier, and Andrew Zisserman.
\newblock A short note on the kinetics-700 human action dataset.
\newblock {\em arXiv preprint arXiv:1907.06987}, 2019.

\bibitem{b39}
Joao Carreira and Andrew Zisserman.
\newblock Quo vadis, action recognition? a new model and the kinetics dataset.
\newblock In {\em proceedings of the IEEE Conference on Computer Vision and
  Pattern Recognition}, pages 6299--6308, 2017.

\bibitem{b27}
Xinyi Cui, Qingshan Liu, Mingchen Gao, and Dimitris~N Metaxas.
\newblock Abnormal detection using interaction energy potentials.
\newblock In {\em CVPR 2011}, pages 3161--3167. IEEE, 2011.

\bibitem{b42}
Jia Deng, Wei Dong, Richard Socher, Li-Jia Li, Kai Li, and Li~Fei-Fei.
\newblock Imagenet: A large-scale hierarchical image database.
\newblock In {\em 2009 IEEE conference on computer vision and pattern
  recognition}, pages 248--255. Ieee, 2009.

\bibitem{b28}
Timothy Hospedales, Shaogang Gong, and Tao Xiang.
\newblock A markov clustering topic model for mining behaviour in video.
\newblock In {\em 2009 IEEE 12th International Conference on Computer Vision},
  pages 1165--1172. IEEE, 2009.

\bibitem{b36}
Andrej Karpathy, George Toderici, Sanketh Shetty, Thomas Leung, Rahul
  Sukthankar, and Li~Fei-Fei.
\newblock Large-scale video classification with convolutional neural networks.
\newblock In {\em Proceedings of the IEEE conference on Computer Vision and
  Pattern Recognition}, pages 1725--1732, 2014.

\bibitem{b31}
Louis Kratz and Ko~Nishino.
\newblock Anomaly detection in extremely crowded scenes using spatio-temporal
  motion pattern models.
\newblock In {\em 2009 IEEE Conference on Computer Vision and Pattern
  Recognition}, pages 1446--1453, 2009.

\bibitem{b32}
Weixin Li, Vijay Mahadevan, and Nuno Vasconcelos.
\newblock Anomaly detection and localization in crowded scenes.
\newblock {\em IEEE Transactions on Pattern Analysis and Machine Intelligence},
  36(1):18--32, 2014.

\bibitem{b11}
Wen Liu, Weixin Luo, Zhengxin Li, Peilin Zhao, Shenghua Gao, et~al.
\newblock Margin learning embedded prediction for video anomaly detection with
  a few anomalies.
\newblock In {\em IJCAI}, pages 3023--3030, 2019.

\bibitem{b17}
Wen Liu, Weixin Luo, Dongze Lian, and Shenghua Gao.
\newblock Future frame prediction for anomaly detection--a new baseline.
\newblock In {\em Proceedings of the IEEE conference on computer vision and
  pattern recognition}, pages 6536--6545, 2018.

\bibitem{b40}
Ze~Liu, Jia Ning, Yue Cao, Yixuan Wei, Zheng Zhang, Stephen Lin, and Han Hu.
\newblock Video swin transformer.
\newblock {\em arXiv preprint arXiv:2106.13230}, 2021.

\bibitem{b3}
Cewu Lu, Jianping Shi, and Jiaya Jia.
\newblock Abnormal event detection at 150 fps in matlab.
\newblock In {\em Proceedings of the IEEE international conference on computer
  vision}, pages 2720--2727, 2013.

\bibitem{b34}
Cewu Lu, Jianping Shi, and Jiaya Jia.
\newblock Abnormal event detection at 150 fps in matlab.
\newblock In {\em Proceedings of the IEEE international conference on computer
  vision}, pages 2720--2727, 2013.

\bibitem{b16}
Weixin Luo, Wen Liu, and Shenghua Gao.
\newblock A revisit of sparse coding based anomaly detection in stacked rnn
  framework.
\newblock In {\em Proceedings of the IEEE international conference on computer
  vision}, pages 341--349, 2017.

\bibitem{b43}
Snehashis Majhi, Ratnakar Dash, and Pankaj~Kumar Sa.
\newblock Temporal pooling in inflated 3dcnn for weakly-supervised video
  anomaly detection.
\newblock In {\em 2020 11th International Conference on Computing,
  Communication and Networking Technologies (ICCCNT)}, pages 1--6. IEEE, 2020.

\bibitem{b30}
Ramin Mehran, Alexis Oyama, and Mubarak Shah.
\newblock Abnormal crowd behavior detection using social force model.
\newblock In {\em 2009 IEEE Conference on Computer Vision and Pattern
  Recognition}, pages 935--942, 2009.

\bibitem{b12}
Guansong Pang, Longbing Cao, Ling Chen, and Huan Liu.
\newblock Learning representations of ultrahigh-dimensional data for random
  distance-based outlier detection.
\newblock In {\em Proceedings of the 24th ACM SIGKDD international conference
  on knowledge discovery \& data mining}, pages 2041--2050, 2018.

\bibitem{b13}
Guansong Pang, Chunhua Shen, and Anton van~den Hengel.
\newblock Deep anomaly detection with deviation networks.
\newblock In {\em Proceedings of the 25th ACM SIGKDD international conference
  on knowledge discovery \& data mining}, pages 353--362, 2019.

\bibitem{b44}
Narinder~Singh Punn and Sonali Agarwal.
\newblock Chs-net: A deep learning approach for hierarchical segmentation of
  covid-19 via ct images.
\newblock {\em Neural Processing Letters}, pages 1--22, 2022.

\bibitem{b45}
Narinder~Singh Punn and Sonali Agarwal.
\newblock Rca-iunet: a residual cross-spatial attention-guided inception u-net
  model for tumor segmentation in breast ultrasound imaging.
\newblock {\em Machine Vision and Applications}, 33(2):1--10, 2022.

\bibitem{b29}
Imran Saleemi, Khurram Shafique, and Mubarak Shah.
\newblock Probabilistic modeling of scene dynamics for applications in visual
  surveillance.
\newblock {\em IEEE transactions on pattern analysis and machine intelligence},
  31(8):1472--1485, 2008.

\bibitem{b5}
Waqas Sultani, Chen Chen, and Mubarak Shah.
\newblock Real-world anomaly detection in surveillance videos.
\newblock In {\em Proceedings of the IEEE conference on computer vision and
  pattern recognition}, pages 6479--6488, 2018.

\bibitem{b7}
Yu~Tian, Guansong Pang, Yuanhong Chen, Rajvinder Singh, Johan~W Verjans, and
  Gustavo Carneiro.
\newblock Weakly-supervised video anomaly detection with contrastive learning
  of long and short-range temporal features.
\newblock 2021.

\bibitem{b37}
Du~Tran, Lubomir Bourdev, Rob Fergus, Lorenzo Torresani, and Manohar Paluri.
\newblock Learning spatiotemporal features with 3d convolutional networks.
\newblock In {\em Proceedings of the IEEE international conference on computer
  vision}, pages 4489--4497, 2015.

\bibitem{b38}
Du~Tran, Lubomir~D Bourdev, Rob Fergus, Lorenzo Torresani, and Manohar Paluri.
\newblock C3d: generic features for video analysis.
\newblock {\em CoRR, abs/1412.0767}, 2(7):8, 2014.

\bibitem{b19}
Shashanka Venkataramanan, Kuan-Chuan Peng, Rajat~Vikram Singh, and Abhijit
  Mahalanobis.
\newblock Attention guided anomaly detection and localization in images.
\newblock {\em arXiv preprint arXiv:1911.08616}, 2019.

\bibitem{b2}
Jiang Wang, Yang Song, Thomas Leung, Chuck Rosenberg, Jingbin Wang, James
  Philbin, Bo~Chen, and Ying Wu.
\newblock Learning fine-grained image similarity with deep ranking.
\newblock In {\em Proceedings of the IEEE conference on computer vision and
  pattern recognition}, pages 1386--1393, 2014.

\bibitem{b8}
Sanghyun Woo, Jongchan Park, Joon-Young Lee, and In~So Kweon.
\newblock Cbam: Convolutional block attention module.
\newblock In {\em Proceedings of the European conference on computer vision
  (ECCV)}, pages 3--19, 2018.

\bibitem{b26}
Shandong Wu, Brian~E. Moore, and Mubarak Shah.
\newblock Chaotic invariants of lagrangian particle trajectories for anomaly
  detection in crowded scenes.
\newblock {\em 2010 IEEE Computer Society Conference on Computer Vision and
  Pattern Recognition}, pages 2054--2060, 2010.

\bibitem{b20}
Dan Xu, Elisa Ricci, Yan Yan, Jingkuan Song, and Nicu Sebe.
\newblock Learning deep representations of appearance and motion for anomalous
  event detection.
\newblock {\em arXiv preprint arXiv:1510.01553}, 2015.

\bibitem{b4}
Bin Zhao, Li~Fei-Fei, and Eric~P. Xing.
\newblock Online detection of unusual events in videos via dynamic sparse
  coding.
\newblock {\em CVPR 2011}, pages 3313--3320, 2011.

\bibitem{b35}
Bin Zhao, Li~Fei-Fei, and Eric~P. Xing.
\newblock Online detection of unusual events in videos via dynamic sparse
  coding.
\newblock {\em CVPR 2011}, pages 3313--3320, 2011.

\bibitem{b6}
Jia-Xing Zhong, Nannan Li, Weijie Kong, Shan Liu, Thomas~H Li, and Ge~Li.
\newblock Graph convolutional label noise cleaner: Train a plug-and-play action
  classifier for anomaly detection.
\newblock In {\em Proceedings of the IEEE/CVF Conference on Computer Vision and
  Pattern Recognition}, pages 1237--1246, 2019.

\bibitem{b21}
Bo~Zong, Qi~Song, Martin~Renqiang Min, Wei Cheng, Cristian Lumezanu, Dae
  ki~Cho, and Haifeng Chen.
\newblock Deep autoencoding gaussian mixture model for unsupervised anomaly
  detection.
\newblock In {\em ICLR}, 2018.

\end{thebibliography}

\end{document}